\documentclass[11pt]{article}

\usepackage{ACL}
\usepackage{float}
\usepackage{xfrac}
\usepackage{times}
\usepackage{latexsym}
\usepackage[T1]{fontenc}
\usepackage[utf8]{inputenc}
\usepackage{microtype}
\usepackage{inconsolata}
\usepackage{hyperref}

\usepackage{multirow}
\usepackage{amsmath}
\usepackage{capt-of}
\usepackage{tabularx}
\usepackage{subcaption}
\usepackage{epsfig}
\usepackage{amssymb}
\usepackage{amsfonts}
\usepackage{booktabs}
\usepackage{scalerel}
\usepackage[inline]{enumitem}
\usepackage{listings}
\usepackage{varwidth}
\usepackage[export]{adjustbox}
\usepackage{tikz}
\usetikzlibrary{tikzmark}

\usepackage{stmaryrd}
\usepackage{bbm}
\usepackage{wrapfig}
\usepackage{pifont}

\definecolor{deepblue}{rgb}{0,0,0.5}
\definecolor{officeblue}{RGB}{0,102,204}
\definecolor{deepred}{rgb}{0.6,0,0}
\definecolor{deepgreen}{rgb}{0,0.5,0}
\definecolor{mybrickred}{RGB}{182,50,28}

\definecolor{fillcolor}{RGB}{216,217,252}

\usepackage{amsmath,amsfonts,bm}

\def\eqref#1{equation~\ref{#1}}

\def\1{\bm{1}}

\DeclareMathAlphabet{\mathsfit}{\encodingdefault}{\sfdefault}{m}{sl}
\SetMathAlphabet{\mathsfit}{bold}{\encodingdefault}{\sfdefault}{bx}{n}

\usepackage{pifont}

\title{In-context Learning and Gradient Descent Revisited}

\author{
Gilad Deutch$^{*}$
\And 
Nadav Magar$^*$\\
~~\\
~~~~~~~~~~~~~~~~~~~~~~~~~~~~~~~~~~~~~~~~The Blavatnik School of Computer Science\\
~~~~~~~~~~~~~~~~~~~~~~~~~~~~~~~~~~~~~~~~Tel Aviv University
\And 
Tomer Bar Natan
\And 
Guy Dar}

\date{}

\begin{document}

\maketitle

\def\thefootnote{*}\footnotetext{Equal contribution}\def\thefootnote{\arabic{footnote}}

\begin{abstract}
In-context learning (ICL) has shown impressive results in few-shot learning tasks, yet its underlying mechanism is still not fully understood. A recent line of work suggests that ICL performs gradient descent (GD)-based optimization implicitly. While appealing, much of the research focuses on simplified settings, where the parameters of a shallow model are optimized. In this work, we revisit evidence for ICL-GD correspondence on realistic NLP tasks and models. 
We find gaps in evaluation, both in terms of problematic metrics and insufficient baselines. We show that surprisingly, even untrained models achieve comparable ICL-GD similarity scores despite not exhibiting ICL.
Next, we explore a major discrepancy in the flow of information throughout the model between ICL and GD, which we term \textit{Layer Causality}. We propose a simple GD-based optimization procedure that respects layer causality, and show it improves similarity scores significantly.
Our code implementation is available at: \url{https://github.com/GiilDe/ft-vs-icl}.


\end{abstract}

\section{Introduction}

In recent years, large language models have shown strong emergent in-context learning abilities \cite{brown20gpt3, wei2022emergent} -- where a pretrained model's performance significantly improves on a task by conditioning the language model on a small set of input-label pairs (demonstrations).
Despite substantial research, the inner workings of ICL remain elusive.
At face value, in-context learning and gradient descent-based finetuning have very little in common.
Nevertheless, a series of recent studies discuss apparent similarities between ICL and gradient descent-based optimization, mostly in synthetic scenarios \cite[][\textit{inter alia}]{pmlr-v202-von-oswald23a, vonoswald2023uncovering, akyürek2023learning, ahn2023transformers}. The claim this body of research aims to make is that ICL can implement implicit GD, using in-context demonstrations as training examples. While most of the synthetic setups concern: (1) restricted transformers, (2) simplified regression tasks, and (3) direct training for ICL -- the work of \citet{dai2023gpt} stands out in its ability to demonstrate an ostensible similarity between ICL and GD optimization in (1) full-fledged transformers, (2) for realistic NLP tasks, (3) naturally occurring in models trained only on causal text generation. We call the hypothesis that ICL mimics finetuning \textit{on the model itself} -- as is analyzed in \citet{dai2023gpt} -- the \textit{strong ICL-GD correspondence}. We will later discuss how this diverges from the ICL-GD correspondence other works consider.

In this paper, we make two main complementary contributions. We perform a careful re-analysis of the work of \citet{dai2023gpt} and show how seemingly mild problems in evaluation lead to a significant overestimation of similarity between the two procedures. Surprisingly, we find that untrained models achieve similarity scores \textit{at least as good} as trained ones. This result provides strong evidence \textit{against} the strong ICL-GD correspondence.\footnote{However, it should be noted that the similarity metrics assume a certain correspondence holds in every layer in a specific way. It does not preclude the possibility of a relaxed correspondence.} 

Secondly, in an attempt to relax the strong ICL-GD correspondence hypothesis, we suggest a rectified version of GD that we show aligns better with ICL. To do this, we first identify a core discrepancy in the flow of information throughout the model between in-context learning and vanilla gradient descent, which we call \emph{Layer Causality}. In ICL, the information that influences the hidden state comes from the output of shallow layers (``earlier layers'') alone. In GD, however, the update to the weights of a layer depends on gradients, which come from all of the model layers including deeper (``later layers''). We showcase the importance of this simple observation by suggesting a simple variant of GD that incorporates layer causality. This simple modification, \textit{Layer Causal Gradient Descent} (LCGD), consistently improves upon vanilla gradient descent on the similarity metrics. Notably, it outperforms the trained transformer significantly in terms of both similarity metrics. In comparison to the untrained baselines, it significantly surpasses them in attention map similarity ($\text{SimAM}_\Delta$) and is consistently on the high end in terms of hidden state similarity (SimAOU). In spite of that, the scores are still low. This can be due to a suboptimal choice of hyperparameters but likely has to do with inherent problems in the strong ICL-GD correspondence hypothesis, even with the layer causal version we propose. We leave this for future work to explore. 

Lastly, we dedicate a short discussion to the line of work on synthetic settings that builds on insights from \citet{pmlr-v202-von-oswald23a}. We observe terminology differences with \citet{dai2023gpt} that might cause confusion. ``Gradient Descent'' is used differently in both cases. While synthetic settings usually consider gradients of shallow implicit functions, \citet{dai2023gpt} consider complex gradients with respect to the model itself. In the synthetic setting, layer causality is often trivially satisfied. 



Our contributions are the following:
\begin{itemize}[label=$\blacktriangleright$]
    \item We discuss issues in the evaluation process of \citet{dai2023gpt} in terms of baselines and evaluation metrics. Notably, we demonstrate that untrained transformers perform as well as pretrained models. 
    \item We highlight core problems with the hypothesis that GD approximates ICL in the naive sense. We study a layer-causal GD variant and demonstrate empirically that it is better at simulating ICL. 
    \item Finally, we briefly survey works in synthetic settings and find that their ICL-GD correspondence is significantly different from the strong ICL-GD correspondence which we try to refute.   
\end{itemize}
In summary, our work shows there's little evidence for the strong ICL-GD correspondence in its current form. We show a non-trivial increase in the similarity metrics (especially in $\text{SimAM}_\Delta$) with a layer-causal variant. This might suggest that a weaker, more nuanced hypothesis might hold. However, we acknowledge there may be irrelevant causes for the increase.

All code for replicating our work is publicly available at:  \href{https://github.com/GiilDe/ft-vs-icl}{https://github.com/GiilDe/ft-vs-icl}.
\begin{table*}[h!]
	\centering
        \hspace*{-0.3cm}
	\setlength{\tabcolsep}{7pt}
		\begin{tabular}{l | l | c c c c c c c }
			\toprule
			& & \textbf{SST2} & \textbf{SST5} & \textbf{MR} & \textbf{Subj} & \textbf{CB}         \\
			 \midrule
              SimAOU & Trained & 0.05\tiny{$\pm$ 0.01} & 0.04\tiny{$\pm$ 0.02} &  0.17\tiny{$\pm$ 0.03} & 0.06\tiny{$\pm$ 0.01}  & \textbf{0.11}\tiny{$\pm$ 0.01}         \\
    		  & Trained Embeddings & \textbf{0.11}\tiny{$\pm$ 0.02}  & \underline{0.06}\tiny{$\pm$ 0.00} & \textbf{0.24}\tiny{$\pm$ 0.00}  & \textbf{0.20}\tiny{$\pm$ 0.00}  & {0.01}\tiny{$\pm$ 0.00} \\
               & No Training & {0.09}\tiny{$\pm$ 0.00}  & \underline{\textbf{0.07}}\tiny{$\pm$ 0.03} & {0.18}\tiny{$\pm$ 0.03}  & {0.06}\tiny{$\pm$ 0.01}  & {0.04}\tiny{$\pm$ 0.01} \\

       \midrule
           $\text{SimAM}_\Delta$ & Trained & \textbf{0.15}\tiny{$\pm$ 0.02}  & \textbf{0.31}\tiny{$\pm$ 0.02} & 0.14\tiny{$\pm$ 0.05}  & \textbf{0.25}\tiny{$\pm$ 0.07}  & \underline{0.25}\tiny{$\pm$ 0.01} \\
              & Trained Embeddings & {0.09}\tiny{$\pm$ 0.02}  & {0.03}\tiny{$\pm$ 0.00} & \textbf{0.18}\tiny{$\pm$ 0.02}  & {0.16}\tiny{$\pm$ 0.02}  & {0.05}\tiny{$\pm$ 0.10} \\
    		 & No Training & {0.11}\tiny{$\pm$ 0.04}  & {0.05}\tiny{$\pm$ 0.03} & {0.16}\tiny{$\pm$ 0.03}  & {0.17}\tiny{$\pm$ 0.03}  & \textbf{\underline{0.26}}\tiny{$\pm$ 0.05} \\

		\end{tabular}
		\caption{SimAOU and SimAM comparison of vanilla GD for trained and untrained transformers. When the difference between the highest and second-highest score in a column is $\leq 0.01$, we underline both scores. 
		}
		\label{tab:untrained}
\end{table*}

\section{Preliminaries}
In this work, we build on the benchmark proposed by \citet{dai2023gpt}. We focus on its setting using the same datasets and examine the same similarity metrics to compare the behavior of ICL and finetuning. This section provides details on the benchmark they use. In the next section, we will address problems in the metrics described below.  


\subsection{Datasets}
Following \citet{dai2023gpt}, we use six datasets for our experiment: \textbf{SST2} \cite{socher-etal-2013-recursive}, \textbf{SST5} \cite{socher-etal-2013-recursive}, \textbf{MR} \cite{10.3115/1219840.1219855} and \textbf{Subj} \cite{10.3115/1218955.1218990} are four datasets for sentiment classification; \textbf{AGNews} \cite{NIPS2015_250cf8b5} is a topic classification dataset; and \textbf{CB} \cite{Marneffe2019TheCI} is used for natural language inference. Data statistics are provided in Table~\ref{tab:data_stats} (Appendix~\ref{appendix:data_stats}).

\subsection{Metric I: SimAOU \textit{Normalized}} 
\label{subsec:simaou}
The first metric quantifies the similarity of two setups (finetuning and in-context learning) in terms of the attention output (AO) vector of each layer. More precisely, we quantify the similarity between the \textit{changes} to the AO vector (changes being the difference from the AO vector in the zero-shot setup). 
Given a test prompt, let $h^{(l)}_S$ be the output representation of the last token at the $l$-th attention layer in setting $S$  where $S \in \{\text{ZSL}, \text{ICL}, \text{FT}\}$ -- zero-shot learning, in-context learning, and finetuning.
The updates induced by ICL and finetuning are given by $h^{(l)}_{\text{ICL}} - h^{(l)}_{\text{ZSL}}$ and $h^{(l)}_{\text{FT}} - h^{(l)}_{\text{ZSL}}$, respectively.
The \textit{attention output update similarity} (SimAOU) is defined as the cosine similarity between these updates, averaged across all layers.
A high SimAOU score indicates that ICL adjusts the attention output in the same direction as finetuning.
As a baseline, they compare with random attention output updates: $h^{(l)}_{\text{rand}} - h^{(l)}_{\text{ZSL}}$ where $h^{(l)}_{\text{rand}}$ is sampled uniformly. We note that the authors used a slight variation of this, where $h^{(l)}_S$ is \textit{normalized} before computing the difference. We call this metric $\text{SimAOU}_\text{norm}$ and would later show that this normalization can cause misleading results. 

\subsection{Metric II: SimAM} 
\label{subsec:simam}
SimAM is used to measure the similarity between attention \textit{maps} of ICL and finetuning.
Given a test example, let $m^{(l,h)}_S$ represent the attention weights before softmax in the $h$-th head of the $l$-th layer for setting $S$. In ICL, we focus solely on the test examples' token attention weights, excluding demonstration tokens so that the shapes of FT and ICL attention weights will be compatible. We calculate the cosine similarity between $m^{(l,h)}_{\text{ICL}}$ and $m^{(l,h)}_{\text{FT}}$ to obtain SimAM. Notice here we do not measure the similarity between \textit{updates} but rather between the raw attention weights themselves. We will return to this shortly when we analyze the metric choices and biases they introduce into the benchmark.

\section{Rethinking the Benchmark}
\subsection{$\text{SimAOU}$}
In the original setting, \citet{dai2023gpt} have shown that random noise gets a minuscule score on this metric. However, we show that even two random update vectors of sufficient norm can achieve a high SimAOU score. Let $\mathbf{z} = h^{(l)}_{\text{ZSL}}$ be the \textit{unnormalized} attention output in zero-shot. Assume $\mathbf{r}, \mathbf{r}' \sim \mathcal{N}\left(0, \sigma I \right)$ are random gaussian noise vectors with variance $\sigma^2$. Now, choose $\sigma$ such that $\|\mathbf{r}\|^2 = \|\mathbf{r}'\|^2 = 3\|\mathbf{z}\|^2$ holds\footnote{This will make the computation cleaner, but other options such as $\|\textbf{z}\| = \|\textbf{r}\| = \|\textbf{r}'\|$ are just as good, leading to slightly different similarity scores.} and set $\mathbf{z}_\text{ICL} = \mathbf{z} + \mathbf{r},\ \mathbf{z}_\text{FT} = \mathbf{z} + \mathbf{r}'$. The random vectors are approximately uncorrelated with each other and with $\mathbf{z}$, that is $\mathbf{z}^T \mathbf{r} = \mathbf{r}^T \mathbf{r}' = \mathbf{z}^T \mathbf{r}' = 0$. By the Pythagorean theorem, $\|\mathbf{z}_\text{ICL}\|^2 = \|\mathbf{z} + \mathbf{r}\|^2 = \|\mathbf{z}\|^2 + \|\mathbf{r}\|^2 =  4 \|\mathbf{z}\|^2 = \|\mathbf{z} + \mathbf{r}'\|^2 = \|\mathbf{z}_\text{FT}\|^2$. So, $\|\mathbf{z}_\text{FT}\| = \|\mathbf{z}_\text{ICL}\| = 2 \|\mathbf{z}\|$. We get that $\text{SimAOU}_\text{norm}$ equals:
\begin{gather*}
\frac{\frac{\mathbf{z}_\text{ICL}}{\|\mathbf{z}_\text{ICL}\|} - \frac{\mathbf{z}}{\|\mathbf{z}\|}}{\left\lVert \frac{\mathbf{z}_\text{ICL}}{\|\mathbf{z}_\text{ICL}\|} - \frac{\mathbf{z}}{\|\mathbf{z}\|} \right\rVert} \cdot \frac{\frac{\mathbf{z}_\text{FT}}{\|\mathbf{z}_\text{FT}\|} - \frac{\mathbf{z}}{\|\mathbf{z}\|}}{\left\lVert \frac{\mathbf{z}_\text{FT}}{\|\mathbf{z}_\text{FT}\|} - \frac{\mathbf{z}}{\|\mathbf{z}\|} \right\rVert} = \\
\frac{\frac{\mathbf{z} + \mathbf{r}}{{2} \|\mathbf{z}\|} - \frac{\mathbf{z}}{\|\mathbf{z}\|}}{\left\lVert \frac{\mathbf{z} + \mathbf{r}}{{2} \|\mathbf{z}\|} - \frac{\mathbf{z}}{\|\mathbf{z}\|} \right\rVert} \cdot \frac{\frac{\mathbf{z} + \mathbf{r}'}{{2} \|\mathbf{z}\|} - \frac{\mathbf{z}}{\|\mathbf{z}\|}}{\left\lVert \frac{\mathbf{z} + \mathbf{r}'}{{2}\|\mathbf{z}\|} - \frac{\mathbf{z}}{\|\mathbf{z}\|} \right\rVert} = \\
\frac{\mathbf{r} - \mathbf{z} }{\left\lVert \mathbf{r} - \mathbf{z} \right\rVert} \cdot \frac{\mathbf{r}' - \mathbf{z} }{\left\lVert \mathbf{r}' - \mathbf{z} \right\rVert} = \frac{||\mathbf{z}||^2 }{2 \left\lVert \mathbf{z} \right\rVert \cdot 2 \left\lVert \mathbf{z} \right\rVert} = \frac{1}{4}
\end{gather*}

The problem our computation reveals is the fact that after normalization, $\mathbf{z}$ terms don't cancel out completely and interact with each other. This is a general problem not limited to random noise. We compare unnormalized SimAOU with $\text{SimAOU}_\text{norm}$ in Table \ref{tab:layer_causal} and show it has a substantial impact on the similarity scores. 

\subsection{$\text{SimAM}_\Delta$}


To better measure the similarity between the updates to the attention maps induced by ICL and FT, we suggest a modified metric, $\text{SimAM}_\Delta$. 
Specifically we compute the cosine similarity between $m^{(l,h)}_{\text{ICL}}-m^{(l,h)}_{\text{ZS}}$ and $m^{(l,h)}_{\text{FT}}-m^{(l,h)}_{\text{ZS}}$, the \textit{update} vectors. The new metric is no longer sensitive to the magnitude of the update vector. In the original setting, the cosine similarity might be dominated by $m^{(l,h)}_{\text{ZS}}$ so a model drifting further during FT from $m^{(l,h)}_{\text{ZS}}$ will be penalized even if the update direction is more similar to ICL's. Update size in general can be manipulated by adjusting the learning rate,\footnote{Though effects are not guaranteed to change linearly with the change learning rate, as learning rate change can often have unpredictable effects.} and so should not be a core feature of the similarity metric. 


\subsection{Untrained Transformer Baseline}
We have discussed problems with metrics. We now turn to baselines. We use untrained models as our baseline. In-context learning is an emergent property attained through pretraining \citep{brown20gpt3}, therefore any similarity between the ``ICL''\footnote{Formally speaking, it shouldn't really qualify as ICL, as the model hasn't attained this capability yet.} setup and the finetuning setup on untrained models cannot be attributed to a learned form of mesa-optimization \citep{hubinger2021risks}. In Table \ref{tab:untrained}, we compare the original model with two baselines: a completely untrained model (\textit{No Training}) and a model where we kept the input and output embeddings (including positional embeddings) and layer norms (\textit{Trained Embeddings}). We find that in terms of SimAOU the untrained baselines slightly exceed vanilla GD. 



\begin{table*}[h!]
	\centering
        \hspace*{-0.45cm}
	\setlength{\tabcolsep}{7pt}
		\begin{tabular}{l| c c c c c c c }
			\toprule
			& \textbf{SST2} & \textbf{SST5} & \textbf{MR} & \textbf{Subj} & \textbf{AGNews} & \textbf{CB} & \textbf{Average}         \\
			\midrule
			 SimAOU{\tiny{norm}} \ (GD)            & 0.11\tiny{$\pm$ 0.00} & 0.09\tiny{$\pm$ 0.01} &  0.22\tiny{$\pm$ 0.01} & 0.18\tiny{$\pm$ 0.02}  & 0.31\tiny{$\pm$ 0.04} & 0.21\tiny{$\pm$ 0.01} & 0.187          \\
			 SimAOU{\tiny{norm}} (LCGD)  & \textbf{0.22}\tiny{$\pm$ 0.01} & \textbf{0.11}\tiny{$\pm$ 0.00} & \textbf{0.33}\tiny{$\pm$ 0.01}& \textbf{0.35}\tiny{$\pm$ 0.01} & \textbf{0.33}\tiny{$\pm$ 0.01} & \textbf{0.34}\tiny{$\pm$ 0.00} & \textbf{0.279} \\
			 \midrule

             SimAOU (GD)            & 0.05\tiny{$\pm$ 0.01} & 0.04\tiny{$\pm$ 0.02} &  0.17\tiny{$\pm$ 0.03} & 0.06\tiny{$\pm$ 0.01}  & \textbf{0.18}\tiny{$\pm$ 0.03} & 0.11\tiny{$\pm$ 0.01} & 0.102          \\
			 SimAOU (LCGD)  & \textbf{0.13}\tiny{$\pm$ 0.01} & \textbf{0.11}\tiny{$\pm$ 0.01} & \textbf{0.21}\tiny{$\pm$ 0.03}& \textbf{0.18}\tiny{$\pm$ 0.00} & 0.13\tiny{$\pm$ 0.01} & \textbf{0.24}\tiny{$\pm$ 0.01} & \textbf{0.167} \\
			 \midrule
			 SimAM (GD)         & \textbf{0.59}\tiny{$\pm$ 0.01}  & \textbf{0.40}\tiny{$\pm$ 0.03} & \textbf{0.49}\tiny{$\pm$ 0.01}  & \textbf{0.45}\tiny{$\pm$ 0.06}  & \textbf{0.48}\tiny{$\pm$ 0.04} & \textbf{0.20}\tiny{$\pm$ 0.03} & \textbf{0.435} \\
			 SimAM (LCGD)     & 0.58\tiny{$\pm$ 0.01} & 0.39\tiny{$\pm$ 0.03} & 0.30\tiny{$\pm$ 0.00} & 0.27\tiny{$\pm$ 0.01} & 0.12\tiny{$\pm$ 0.00} & 0.04\tiny{$\pm$ 0.01} & 0.283          \\
              \midrule
			 $\text{SimAM}_\Delta$ (GD) & 0.15\tiny{$\pm$ 0.02}  & 0.31\tiny{$\pm$ 0.02} & 0.14\tiny{$\pm$ 0.05}  & 0.25\tiny{$\pm$ 0.07}  & \textbf{0.50}\tiny{$\pm$ 0.05} & 0.25\tiny{$\pm$ 0.01} & 0.267 \\
			 $\text{SimAM}_\Delta$ (LCGD) & \textbf{0.30}\tiny{$\pm$ 0.02}  & \textbf{0.33}\tiny{$\pm$ 0.01} & \textbf{0.26}\tiny{$\pm$ 0.00}  & \textbf{0.32}\tiny{$\pm$ 0.01}  & 0.43\tiny{$\pm$ 0.02} & \textbf{0.38}\tiny{$\pm$ 0.01} &  \textbf{0.336} \\
		\end{tabular}
		\caption{SimAOU and SimAM comparison of vanilla GD and layer-causal GD across six classification datasets.
			Layer causal GD achieves higher SimAOU across all tasks, yet its SimAM is significantly lower. $\text{SimAM}_\Delta$ is higher for layer causal GD, except for AGNews.
		}
		\label{tab:layer_causal}
\end{table*}
		
\begin{figure}[h!]%
	\centering
	\subfloat{{
 \hspace*{-0.3cm}
 \includegraphics[width=7.8cm]{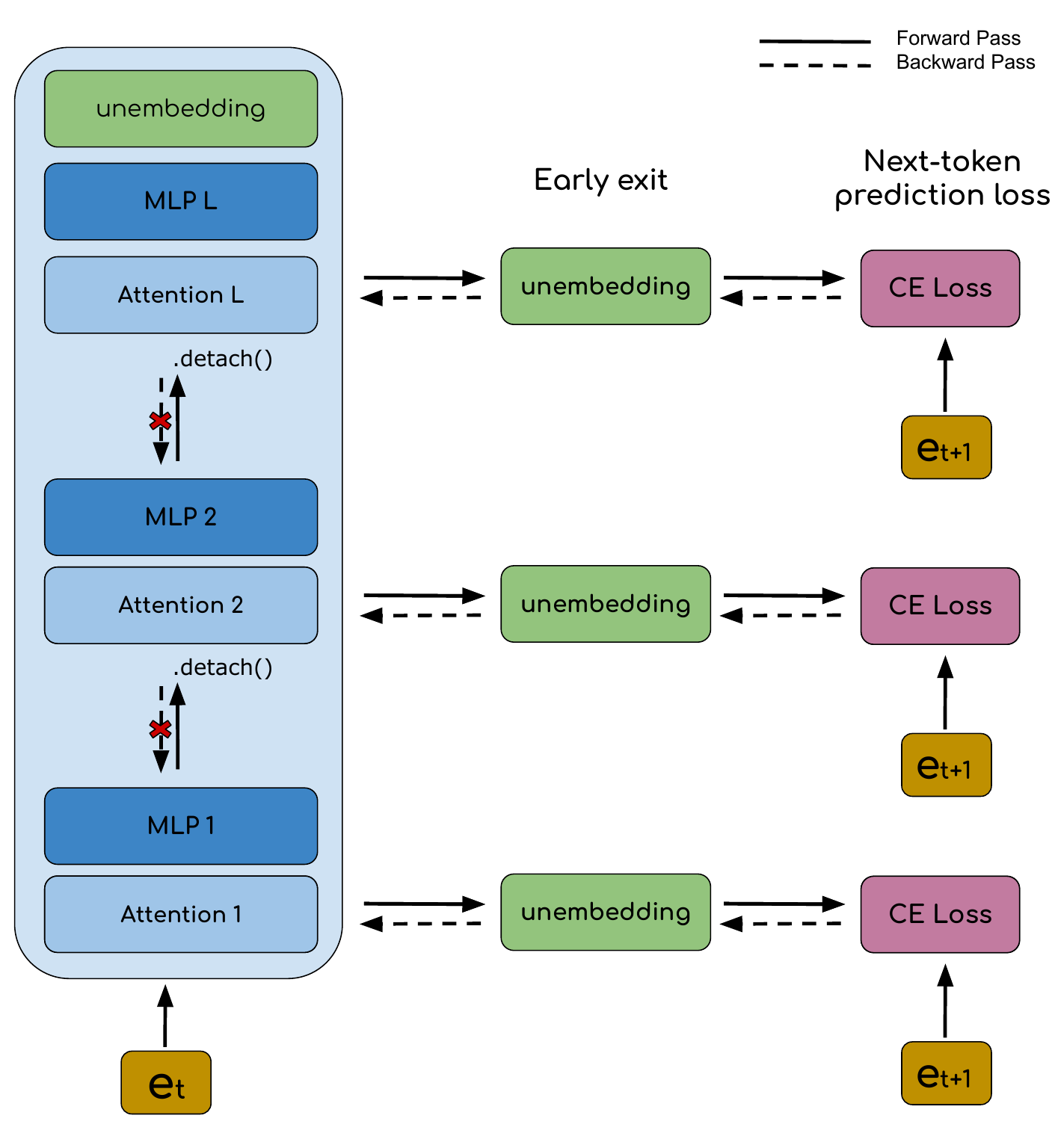}}}%
	\caption{\textbf{Layer-causal GD}: The output of each layer is projected to the label space and used as an intermediate prediction.
		We compute the prediction loss of each intermediate layer sequentially.
	}
	\label{fig:layer_causal_diagram}
\end{figure}

\section{Investigation into Layer Causality}
\label{sec:layer_causality}
\subsection{Layer Causality}
We characterize a core problem with the strong ICL-GD correspondence in the following statement.
\linebreak

\noindent \emph{\textbf{Layer Causality}}. \ \textit{In ICL, the update to the output of the $l$-th attention layer is dependent only on the output of previous (lower) layers.
  In contrast, the update to the $l$-th attention output induced by finetuning is determined by the gradient of the entire model's trainable parameters.}
\linebreak

	

\subsection{Design Choices}
Motivated by this observation, we propose to use a layer causality-compatible finetuning method, where each layer is updated individually, instead of propagating information back to earlier layers. Then, we will explore how a layer-causal variant fares compared to full-blown vanilla gradient descent. There are many possible ways to design such an algorithm. In this work, we will define an instantiation of layer causality-compatible optimization, that we call \textit{Layer-causal Gradient Descent} (LCGD). We make the decision based on the following guiding principles:
\begin{itemize}[label=$\triangleright$]
    \item \textbf{Minimal Changes}: We want to leave the procedure as close as possible to vanilla GD. The goal is to isolate the effect of layer causality on the modification we make as much as possible. Otherwise, other design decisions might come into play.  
    \item \textbf{Simplicity}: We want the procedure to be interpretable and easy to reason about. 
    \item \textbf{Plausibility} (Occam's razor): We want to design a ``plausible'' procedure. A major part of what we call plausibility is layer causality. Plausibility in a broader sense may include any other aspect that one cannot expect a forward pass of the model to easily implement using a clear and simple mechanism. 
\end{itemize}
These principles might conflict. We prioritize them in the following way: we want the procedure to be layer-causal (a special case of plausibility), but other than that, we will always favor the first and second principles.  One example of where we favor simplicity over plausibility is when we choose to take the derivative of the entire layer on every step of the procedure (see below), including the $\operatorname{softmax}$ operation. This goes against plausibility because the derivative of $\operatorname{softmax}$ cannot be plausibly computed with a single attention layer. 

\subsection{Motivation: Short-circuited Transformers}
A simple finetuning method that respects layer causality is by short-circuiting a model at any layer $l$, i.e. by removing all layers from $l+1$ onwards. In a normal (not short-circuited) forward pass, the model outputs the next-token prediction by taking the final hidden state, applying a final layer norm operation to it, and multiplying by the output embedding matrix (a.k.a. the unembedding matrix).  Analogously, in a model short-circuited at layer $l$, the next-token prediction is obtained by projecting the $l$-th hidden state on the unembedding matrix, after applying the final layer norm. This is justified by the \textit{early exit} approach \cite{early_exit, din2023jump}, where it has been observed that a short-circuited model is often sufficiently good at predicting the next token. Early exit is closely related to the residual stream hypothesis \cite{nostalgebraist, elhage2021mathematical, geva2022transformer, dar2023analyzing}, which stipulates that language models refine the next-token prediction throughout the layers -- and so projecting internal layers into the vocabulary space gives the current prediction in every layer.
We will refer to the combination of the final layer norm and the unembedding matrix as the unembedding projection head and denote it by the function $U(\cdot)$.

\subsection{Algorithm}
We now describe the LCGD finetuning procedure. In LCGD we project the output of each layer onto logits in the vocabulary space using the unembedding head $U(\cdot)$ and compute the cross-entropy loss of this prediction with respect to the one-hot embedding of the next token. Unlike vanilla finetuning, it does not violate the causal structure of the network, as it depends only on data available at this layer. To reiterate, $U(\cdot)$ normally takes the final hidden state of the model and projects it onto the logits over the vocabulary. We follow the early exit/residual stream approach and apply it on internal hidden states.

Let the \textit{detached} hidden states after the $\ell$-th attention layer at token $i$ be denoted:
\[
\hat{h}_i^\ell = \operatorname{Attn}\left(W_V \textsc{SG}(X^\ell), W_K\textsc{SG}(X^\ell), \textsc{SG}(\mathbf{q}_i^\ell)\right) 
\] 
where $\operatorname{SG}(\cdot)$ stands for the ``stop gradient'' operation (also called \texttt{.detach()} in PyTorch) which does not affect the forward pass, but in the backward pass it does not back-propagate the gradient to its input, meaning it is treated as a constant.
Let the tokens of the model be represented by a list of one-hot vectors $\mathbf{e}_1, \mathbf{e}_2, ..., \mathbf{e}_T$. For each token, we define the objective function:\footnote{Notice that we are allowed to take the sum of the cross-entropies across all layers in parallel, as the updates to the weight matrices will take effect only when processing the next token.}
\begin{equation*}
\begin{aligned}
\label{eq:loss}
\mathcal{L} = & \sum_{\ell=1}^{L} \textrm{CE}\left(U(\hat{h}_i^\ell), \mathbf{e}_{i+1}\right) 
\end{aligned}
\end{equation*}
$U$ is taken to be frozen as well. $\textrm{CE}$ is cross-entropy loss. 
We optimize by taking steps with respect to the gradient $\nabla_W \mathcal{L}$, one token at a time, where the ``stop gradient'' operator makes sure each layer is updated independently. 


\subsection{Experimental Setup}
\label{sec:exp_analysis}
We use the same GPT-like pre-trained language models used by \citet{dai2023gpt} with 1.3B implemented in fairseq.\footnote{\url{https://github.com/facebookresearch/fairseq}} We test vanilla and layer-causal GD in terms of their similarity to ICL with the four variants we discussed above (SimAOU, $\text{SimAOU}_\text{norm}$, SimAM, $\text{SimAM}_\Delta$). For reliable results, we average across 3 different seeds. This whole project's computation took the equivalent of 12 hours on a single Tesla V100 GPU. Table~\ref{tab:layer_causal} shows both variants of SimAOU and SimAM for both methods.

Overall, with the exception of \textbf{AGNews}, layer-causal GD is significantly more aligned with ICL in terms of the modified similarity metrics and the normalized variant of SimAOU. However, it is important to note that the modified metrics are low for both variants. In comparison to untrained transformers, LCGD is much better in terms of $\text{SimAM}_\Delta$ and is mostly better by some small margin in terms of SimAOU.

\paragraph{Comparison with Untrained Baselines} Combining Tables \ref{tab:untrained} \& \ref{tab:layer_causal}, we see that LCGD is competitive with respect to all three contenders, showing high-end scores consistently across the board, while it is not always the highest in terms of SimAOU. In terms of $\text{SimAM}_\Delta$ is \textit{significantly} better than any of the other baselines across all datasets explored. There remains work to be done to show this advantage is indeed due to structural superiority and not rudimentary features, such as its ability to impact layers more strongly (as the gradient norm of updates in LCGD is larger -- see Appendix~\ref{appendix:grad_norm}), which could have accumulating effects across layers and timesteps. Even if this is the case, it is important to understand the implications of this observation on other variants as well. We leave it for future research to work out the correct interpretation of the results in this section.

\subsection{Additional Experiments} 

\begin{figure*}
	\centering
	{\includegraphics[height=6.5cm]{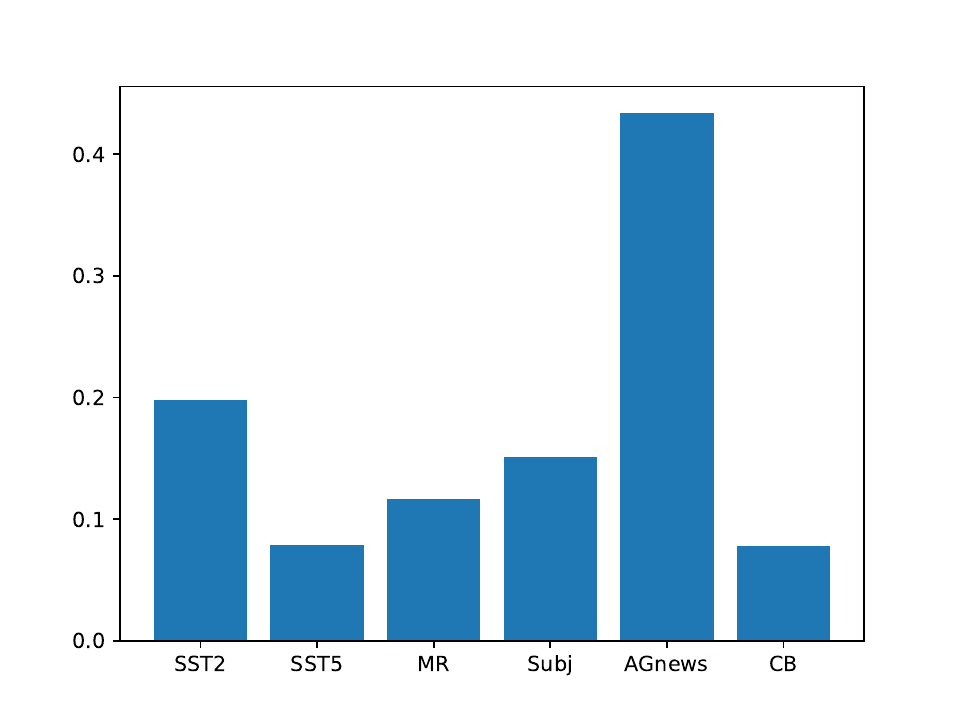}}
	\caption{$\alpha$ averaged over all layers for each task. Computed for one seed per task.}
	\label{fig:causal_vs_gd_cos}
\end{figure*}

\begin{figure*}
	\centering
        \vspace*{-0.25cm}
	{\includegraphics[width=0.91\textwidth]{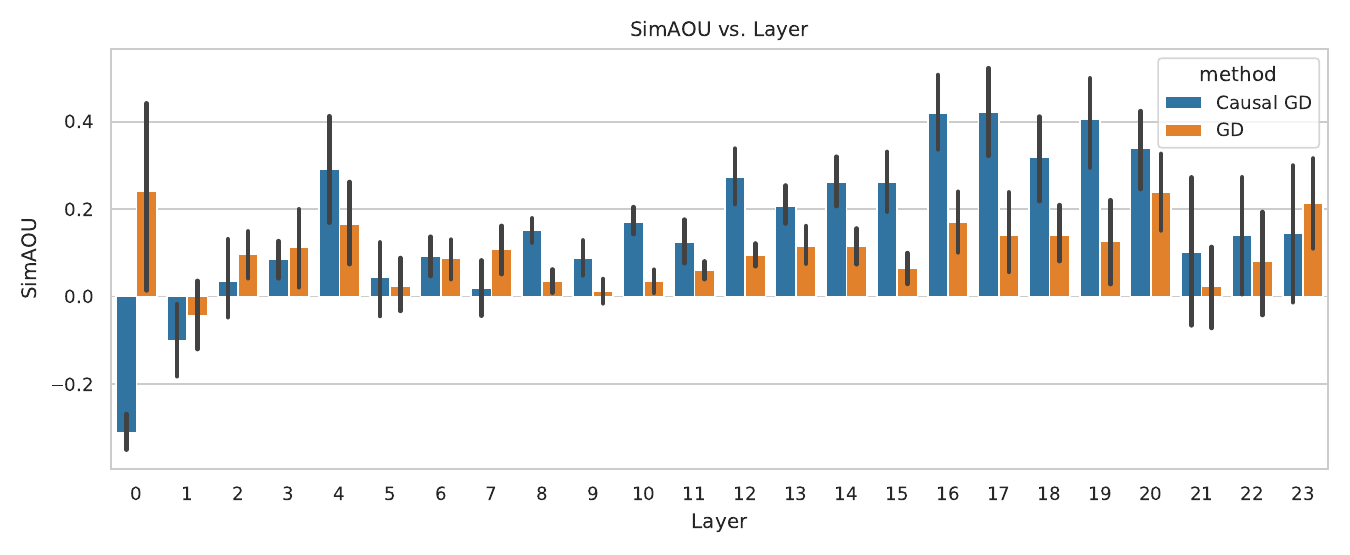}}
	{\includegraphics[width=0.91\textwidth]{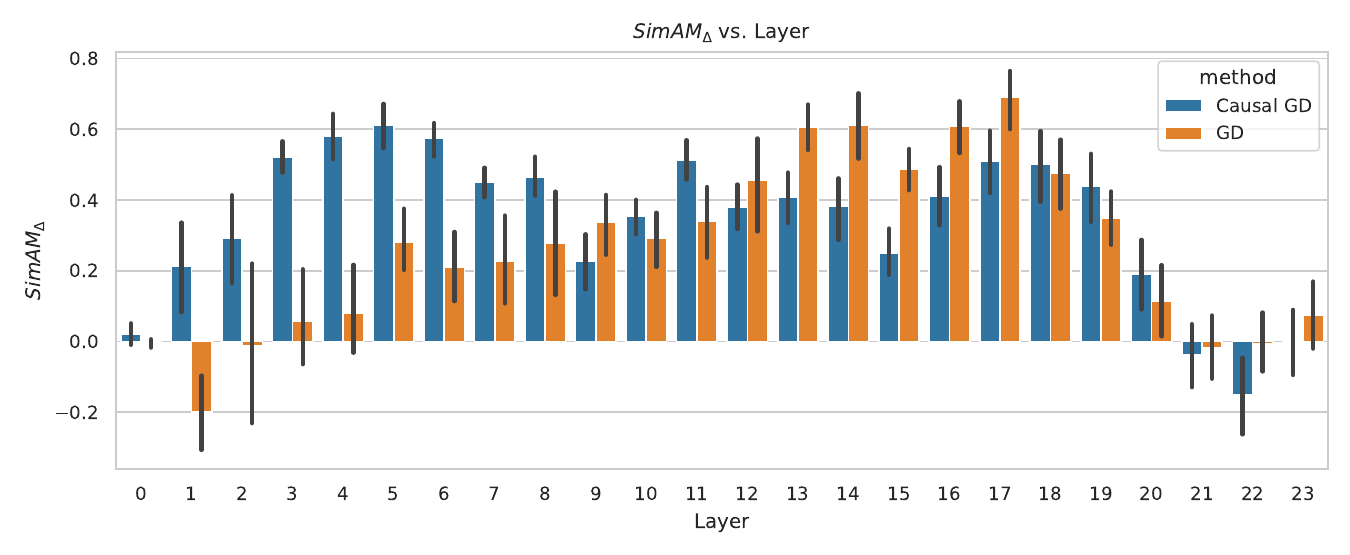}}
	\caption{Similarity computed per layer aggregated across tasks and seeds. Error bar is presented. Blue bars represent layer causal GD and orange is used for vanilla GD. \textit{Top}: SimAOU of each layer's update vector. \textit{Bottom}: $\text{SimAM}_\Delta$ of each layer's update vector.}
	\label{fig:layerwise_sim}
\end{figure*}

In Appendix~\ref{appendix:layer_causal_delve}, we perform a more fine-grained comparison of LCGD and vanilla GD. First, we try to assess how similar the two variants are in the latent space, the intuition being that the layer-causal variant can be a simple approximation to vanilla GD. We find that this similarity is in fact relatively low, around 0.1 more or less in terms of cosine similarity, across datasets (this is shown in Figure~\ref{fig:causal_vs_gd_cos}). Then, we perform a layerwise analysis of the way the similarity scores change.  The results are shown in Figure~\ref{fig:layerwise_sim}. We see a non-trivial variability in the similarity across layers, which seems to suggest a non-uniform behavior across layers. Curiously, we see that LCGD is not better in all layers. In the case of SimAOU, we see a small advantage for LCGD across virtually all layers, but the dynamics of $\text{SimAM}_\Delta$ are more complicated, suggesting deeper analysis is required to fully understand the advantage of LCGD over GD (see Appendix for more details on the additional experiments). 



\section{Conflation of Terms in ICL-GD Correspondence}
Works rooted in the work of \citet{pmlr-v202-von-oswald23a} usually have a common structure: The model is given training examples of the form $\{(\mathbf{x}_1, y_1), (\mathbf{x}_2, y_2), ..., (\mathbf{x}_n, y_n)\}$, where it holds that $y_i = f_\theta(\mathbf{x}_i)$ for some latent parameter vector $\theta$.\footnote{$f_\theta$ can be stochastic.} The model is also fed a test query $\mathbf{x}_\text{test}$. It is trained to output the value $y_\text{test} = f_\theta(\mathbf{x}_\text{test})$. The function $f_\theta$ is always a shallow function, usually a linear model $f_\theta(\mathbf{x}) = \theta^\top \mathbf{x}$, or a kernel regression problem. This distinction is important since the gradient of such functions has a simple closed form. This is in stark contrast to \citet{dai2023gpt}, where the gradient is unwieldily complicated. Another difference is that the gradient in \citet{dai2023gpt} is computed with respect to the transformer itself, not a subsidiary function $f_\theta$. In these crucial aspects, the gradients discussed are extremely different. The strong ICL-GD correspondence explored in \citet{dai2023gpt} is different than the one that the ICL-shallow GD correspondence \citet{pmlr-v202-von-oswald23a} considered -- the use of the term ``Gradient Descent'' in these two cases is incompatible. In Appendix~\ref{appendix:oswald_works}, we go over a subset of these works to demonstrate what kinds of shallow GD they rely on.

\section{Discussion}
In this work, we provide different perspectives on the ICL-GD correspondence. We show evidence against it but also show that it might be fixed. We find that previous work does not justify the strong ICL-GD correspondence, and instead discusses a weaker notion of a shallow GD. This should also apply to layer-causal GD, as it is designed as a modification of the strong ICL-GD correspondence. Still, we see it outperforms untrained transformers in terms of attention map similarity (and fares well in terms of hidden state similarity). This can be due to irrelevant causes (see limitations below). However, it is worth noting that the layer-causal variant can be justified by its similarity to the kernel regression and functional GD variants that have been addressed in the literature on synthetic settings \citep{cheng2024transformers}. Future work can use the (corrected) similarity metrics suggested in \citet{dai2023gpt} to gauge the similarity of shallow GD methods to ICL.

\section{Limitations}
\begin{itemize}[label=$\triangleright$]
\item \emph{Similarity Metrics}: The similarity metrics we use only consider a very specific correspondence between ICL and GD, where each layer applies GD to the model. However, it is possible that the exact mechanism is different (e.g. not all layers do GD).

\item \emph{Datasets}: We use the same datasets used in the original paper by \citet{dai2023gpt} to make sure we do not introduce factors that benefit our method inadvertently. The dataset collection needs to be diversified. Four out of six datasets are sentiment classification datasets. One of the other tasks, CB, is very small, contributing to instability. Similarly, we consider a specific model in all our experiments. To make a more general claim, other models should be tested too. 

\item \emph{LCGD}: We propose a specific instantiation of layer-causal gradient descent. Better instances may exist. While the results for LCGD are (mildly) encouraging, we were unable to rule out the intervention of different secondary effects in score improvement. Despite our best efforts, we suspect such effects might have taken place. One immediate direction for future work is doing hyperparameter search to understand whether there's an impact of different learning rates on the similarity scores. 
\end{itemize}

\section{Related Work}
\label{sec:related}
Many works consider synthetic settings \cite{akyürek2023learning, pmlr-v202-von-oswald23a, vonoswald2023uncovering, ahn2023transformers, cheng2024transformers}. They are mostly concerned with ICL implementing GD of a shallow model, mostly variants of linear models or kernel regression. 

Unlike these works,  \citet{dai2023gpt}, which we are heavily influenced by, study large GPT transformers on structured language classification tasks. Gradient Descent in \citet{dai2023gpt} is with respect to the transformer itself, which is also a significant departure. \citet{Panigrahi2023TrainableTI} show how a transformer can implement the backward pass of another (smaller) transformer in its forward pass. As far as we know, there is no indication that this process is happening in real-world models. 

Recently, new works have emerged \cite{todd2023function, hendel2023incontext} suggesting a different approach to interpreting ICL as an algorithm that compresses training demonstrations into a function/task vector that steers the model to perform the task. Other perspectives of ICL include induction heads \cite{Olsson2022IncontextLA} and Bayesian inference \citep{xie2022explanation}. 

The work of \citet{shen2023pretrained} points to another discrepancy between \textit{full-batch} GD and ICL. They show that vanilla full-batch GD and ICL cannot be reconciled due to ICL's sensitivity to the order of the demonstrations, while full-batch GD is invariant to it. However, this discrepancy can be mitigated easily by applying GD sequentially, as was done in the work of \citet{dai2023gpt} that we compare to. 

Layer causal GD is similar to \citet{bengio2006greedy}, where a similar idea was proposed to accelerate training by finding a good starting point using a greedy layer-wise approach. 
\section{Conclusions}

Inspired by recent work, we explore the relationship between in-context learning and gradient descent-based finetuning in practical settings. We show problems with the strong version of the ICL-GD correspondence. We correct the similarity metrics used in prior work and propose alternatives. Furthermore, we show that a simple baseline of untrained models has higher similarity scores compared to trained models. Our work suggests considering the possibility that only a weak version of ICL-GD holds. We rely on layer causality to further justify this view. We study a potential workaround to this problem (LCGD) that does not violate layer causality and get mixed results. The study of LCGD is not comprehensive enough to make a definite statement for or against layer-causal GD mesa-optimizers. We note a potential connection to kernel regression and functional GD, that come up in works on synthetic setups that uphold the weak ICL-GD correspondence. We leave for future work to elucidate the nature of these connections, as well as propose better layer-causal variants.
\bibliography{refs}

\newpage
\appendix
{
\onecolumn 
\section{Data Statistics}
\label{appendix:data_stats}
\begin{table}[H]
	\centering
	\setlength{\tabcolsep}{7pt}
		\begin{tabular}{l | c | c | c }
			\toprule
			&   \# Train & \# Validation & Avg. \# of Tokens\\
			 \midrule
              \textbf{SST2} & 67,349 & 1,821 & 55.43 \\
    		 \textbf{SST5} & 8,544 & 2,210 & 102.95 \\
               \textbf{MR} &  8,530 & 1,066 & 113.39 \\
                \textbf{Subj}  & 8,000 & 2,000 & 129.23 \\
                \textbf{AGNews}  & 120,000 & 7,600 & 237.72 \\
                \textbf{CB}  & 250 & 250 & 295.80
		\end{tabular}
		\caption{Data statistics of all the datasets in the benchmark}
		\label{tab:data_stats}
\end{table}

\section{Deeper Analysis of Layer Causality}
\label{appendix:layer_causal_delve}
\subsection{Does Layer Causal Gradient Descent Approximate Gradient Descent?}
A natural question that might arise is how similar GD is to the suggested layer causal method. Due to their relatively similar scores, one might conjecture that layer causal GD is a low-resource approximation for GD. We can gauge how similar the two update vectors are to each other using a variant of the attention map metric: $\text{SimAM}^{\text{GD},\ \text{LCGD}}_\Delta =
\text{CosSim}\left(m^{(l,h)}_{\text{LCGD}}-m^{(l,h)}_{\text{ZS}}, m^{(l,h)}_{\text{GD}}-m^{(l,h)}_{\text{ZS}}\right)$. 
This way we can measure how much of the score is attributable to the similarity between the update vectors. We will denote the metric by $\alpha$.


We take one seed per task and compute the average $\alpha$ over the layers, for each task. Counter to our expectations, Figure~\ref{fig:causal_vs_gd_cos} shows that for most datasets, $\alpha \approx 0.1-0.2$, which is very low. This shows that the updates are \textit{not} very correlated, and most of the score of either of the procedures cannot be attributed to a common direction in space.

\subsection{Layerwise Analysis}
Until now, the metrics reported are averaged across all layers. However, it is interesting to look at similarity patterns across layers. In Figure~\ref{fig:layerwise_sim}, we show the SimAOU and $\text{SimAM}_\Delta$ scores averaged across all tasks and seeds for each layer. Interesting patterns emerge in the plots. First, we notice that LCGD outperforms vanilla GD in terms of SimAOU (except for layers 1, 3, and the last layer). In the second plot, we have a more complicated case. In the first half of the model, $\text{SimAM}_\Delta$ is greater for the causal variant (except for layer 9). However, for all layers 12-17, vanilla GD is substantially greater than layer causal. Beginning from layer 18, both scores decrease more or less together. 

With this discrepancy between the metrics, it is worth discussing their different roles. SimAOU captures the similarity to ICL's \textit{hidden states}. They have a direct effect on the model's prediction. Attention logits on the other hand only modulate the coefficients that determine the hidden state. The hidden state mediates their interactions with the rest of the model. They have no direct effect on the prediction, conditioned on the hidden state. On the other hand, attention maps can provide us insight into the way attention has shifted as a response to the parameter update. The higher this metric is, the better it replicates the way ICL attends to its input. While not directly affecting the output, it focuses on what ``interests'' ICL.

Finally, it is important to remember that our GD variant was selected intentionally due to its simplicity. Mild modifications might make it a better contender. Moreover, the setting we consider is limited to the one chosen by \citet{dai2023gpt}, including reusing the same hyperparameters for both methods. It is possible that tuning the hyperparameters for our variant would have yielded better results. All in all, we can state rather confidently that even this simple baseline \textit{performs on par with vanilla GD} across multiple benchmarks, and in some cases outperforms it. Furthermore, it has appealing features, such as being low resource, simple, and causally plausible. 

\section{Gradient Norm in LCGD}
\label{appendix:grad_norm}

\begin{figure}[H]
	\centering
	\subfloat{{\includegraphics[width=8cm]{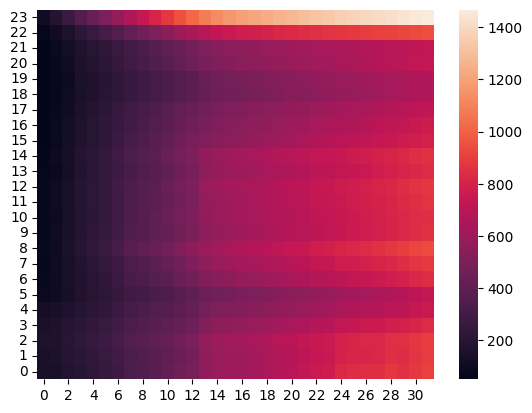}}}%
	\subfloat{{\includegraphics[width=8cm]{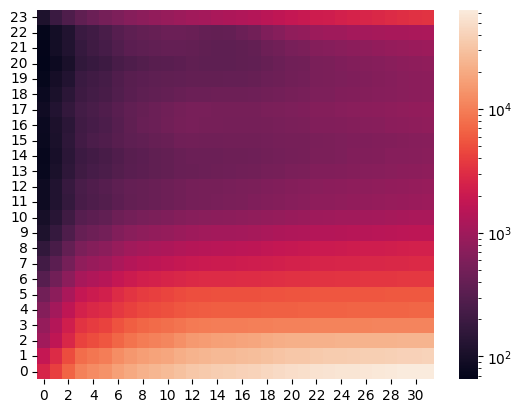}}}%
	\caption{Heatmap of $\ell_2$ norms of the gradients computed during finetuning on the \textbf{Subj} task. Note the different scales of magnitude.
	 \textbf{Horizontal Axis}: Training demonstration index. 
	 \textbf{Vertical Axis}: Layer index in ascending order (from input to network output).
	 \textbf{Left}: Vanilla GD.
	 \textbf{Right}: LCGD (norm magnitude in \textit{logarithmic scale}).
	 }
	\label{fig:grad_norm_comparison}
\end{figure}

\section{Overview of Select Works in the Synthetic Line of Work}
\label{appendix:oswald_works}
\begin{itemize}[label=$\circ$]
    \item \citet{pmlr-v202-von-oswald23a} study linear transformers with data of the form $f_\theta(\mathbf{x}) = \theta^\top \mathbf{x}$. They found a variant of GD (w.r.t. $f_\theta$) that they called $\text{GD}^{++}$ that seems to be implemented by ICL.
    \item \citet{ahn2023transformers} discuss the same linear data scenario. They conclude the optimality of a preconditioned variant of GD/$\text{GD}^{++}$ under different assumptions. 
    \item \citet{vonoswald2023uncovering} study auto-regressive linear transformers. The function under consideration adds stochasticity to the model: $f_W(\mathbf{x}) = W \mathbf{x} + \epsilon$ with $W$ being a matrix instead of a vector, and the input of each demonstration being the previous demonstration. They uncover an intriguing algorithm performed by the transformer, combining preconditioning and GD. 
    \item \citet{cheng2024transformers} discuss transformers with non-linear attention of the form $\mathcal{K}(\mathbf{u}, \mathbf{v})$ where $\mathcal{K}$ is a kernel function. The data in their case comes from a generalized Gaussian process. They consider the empirical quadratic loss objective:
    \begin{equation*}
        \mathcal{L}(f) = \sum_{i=1}^N (f_\theta(\mathbf{x}_i) - {y}_i)^2
    \end{equation*}
    This objective function is more complicated than in other cases described here, as $f_\theta$ is no longer linear. However, they show optimality of gradient descent in \textit{function space}, which turns out to take on a simple form: $\nabla_f \mathcal{L}(f) = \sum_{i=1}^N  (y_i - f(\mathbf{x}_i)) \mathcal{K}(\mathbf{x}_i, \cdot)$. This is in line with the intuition that detached forms of GD are the ones that we should consider, the same intuition as in the construction of layer-causal GD.    
\end{itemize}

\end{document}